\documentclass[letterpaper, 10 pt, conference]{ieeeconf}

\IEEEoverridecommandlockouts                              % This command is only needed if  you want to use the \thanks command

\usepackage[utf8]{inputenc}

\usepackage{graphics} % for pdf, bitmapped graphics files
\usepackage{epsfig} % for postscript graphics files
\usepackage{times} % assumes new font selection scheme installed
\usepackage{amsmath} % assumes amsmath package installed
\usepackage{amssymb}  % assumes amsmath package installed
\usepackage{multirow}
\usepackage{diagbox}
\usepackage{svg}
\usepackage{caption}
\usepackage{tikz}
\usepackage{subcaption}

\usepackage{cite}
\usepackage[noend]{algpseudocode}
\usepackage[ruled,vlined,linesnumbered]{algorithm2e}
\usepackage{algpseudocode}

\usepackage{color}
\usepackage{tikz}
\usepackage{comment}
\usepackage{hyperref}
\usepackage{booktabs}
\usepackage{adjustbox}

\newcommand{\ourmethod}{C-Uniform}
\newcommand{\ourmethody}{C-Uniformity}

\newcommand{\algoname}{\emph{\ourmethod}}
\newcommand{\statevec}{\mathbf{x}}

\newtheorem{problem}{Problem}
\newtheorem{lemma}{Lemma}

\title{\ourmethod{} Trajectory Sampling For Fast Motion Planning}
\author{O. Goktug Poyrazoglu, Yukang Cao and Volkan Isler
}

\date{March 2024}

\begin{document}

\maketitle

\begin{abstract}
We study the problem of sampling robot trajectories and introduce the notion of \ourmethody{}. 
As opposed to the standard method of uniformly sampling control inputs (which lead to biased samples of the configuration space), \ourmethod{} trajectories are generated by control actions which lead to uniform sampling of the configuration space. After presenting an intuitive closed-form solution to generate \ourmethod{} trajectories for the 1D random-walker, we present a network-flow based optimization method to precompute \ourmethod{} trajectories for general robot systems. We apply the notion of \ourmethody{} to the design of Model Predictive Path Integral controllers. Through simulation experiments, we show that using \ourmethod{} trajectories significantly improves the performance of MPPI-style controllers, achieving up to 40\% coverage performance gain compared to the best baseline. We demonstrate the practical applicability of our method with an implementation on a 1/10th scale racer. 
\end{abstract}

% OUTLINE 
% \input{sections/outline}

\section{Introduction}  \label{sec:introduction}
% \paragraph{Why is trajectory sampling important?} Many motion planning algorithms require sampling trajectories from the set of feasible trajectories. Examples include PRMs, RRTs, MPC variants such as MPPI. Trajectory sampling also plays an important role in reinforcement learning. For example, during the exploration stage, ``random" trajectories are generated to estimate the value function. 

% \paragraph{Why is uniform sampling important?}

% We can use a set of sampled trajectories to estimate a value (average cost, expected risk) or to find a trajectory with close to an optimal value.
% We can also use samples to estimate the gradient of a function of the trajectory e.g. for gradient descent.

% In such cases, uniformly sampling the domain (e.g. the space of all trajectories) gives us an unbiased estimator.

% \paragraph{What does uniform sampling mean in our case?}

% The classical definition of uniformity

% %% note to self and also to Yukang

% % 1. The sample mean from uniform sampling is an unbiased estimator of the population mean.

% % 2. Relationship to KL divergence

% % 3. Relationship to the earthmovers theorem

% % Difference between state space sampling and action space sampling. Figure 1. ?

Trajectory sampling is the task of generating ``random" robot trajectories from the set of all possible robot trajectories. Trajectory sampling plays a critical role in randomized motion planning~\cite{orthey_sampling-based_2024}, model predictive control~\cite{williams_information_2017-1} and reinforcement learning~\cite{ota_trajectory_2019}. In this paper, we consider basic, but surprisingly understudied questions related to trajectory sampling: what is a desired goal distribution for sampling trajectories? How can we generate samples according to this distribution? 

To set the stage for our study, consider the most common way of generating trajectory samples: by sampling robot control inputs according to either uniform or Gaussian probability densities. This sampling strategy generates strategies which are helpful for understanding where the robot would be if random disturbances were applied to the input at each time step. But they are not as helpful for generating random trajectories to cover the robot's  configuration space (C-space). As an alternative approach, one might instead directly sample the C-Space but turning these samples into trajectories is not straightforward (it is widely studied in the context of probabilistic roadmaps~\cite{kavraki1996probabilistic},~\cite{karaman_sampling-based_2011}). More importantly, the distribution of the trajectories generated by this approach would be unclear. 

\begin{figure}[t]
    \captionsetup{aboveskip=10pt, belowskip=-15pt} % Set specific spacing for this figure
    \centering
    \begin{subfigure}[b]{\columnwidth}
        \centering
        \includegraphics[width=0.7\textwidth]{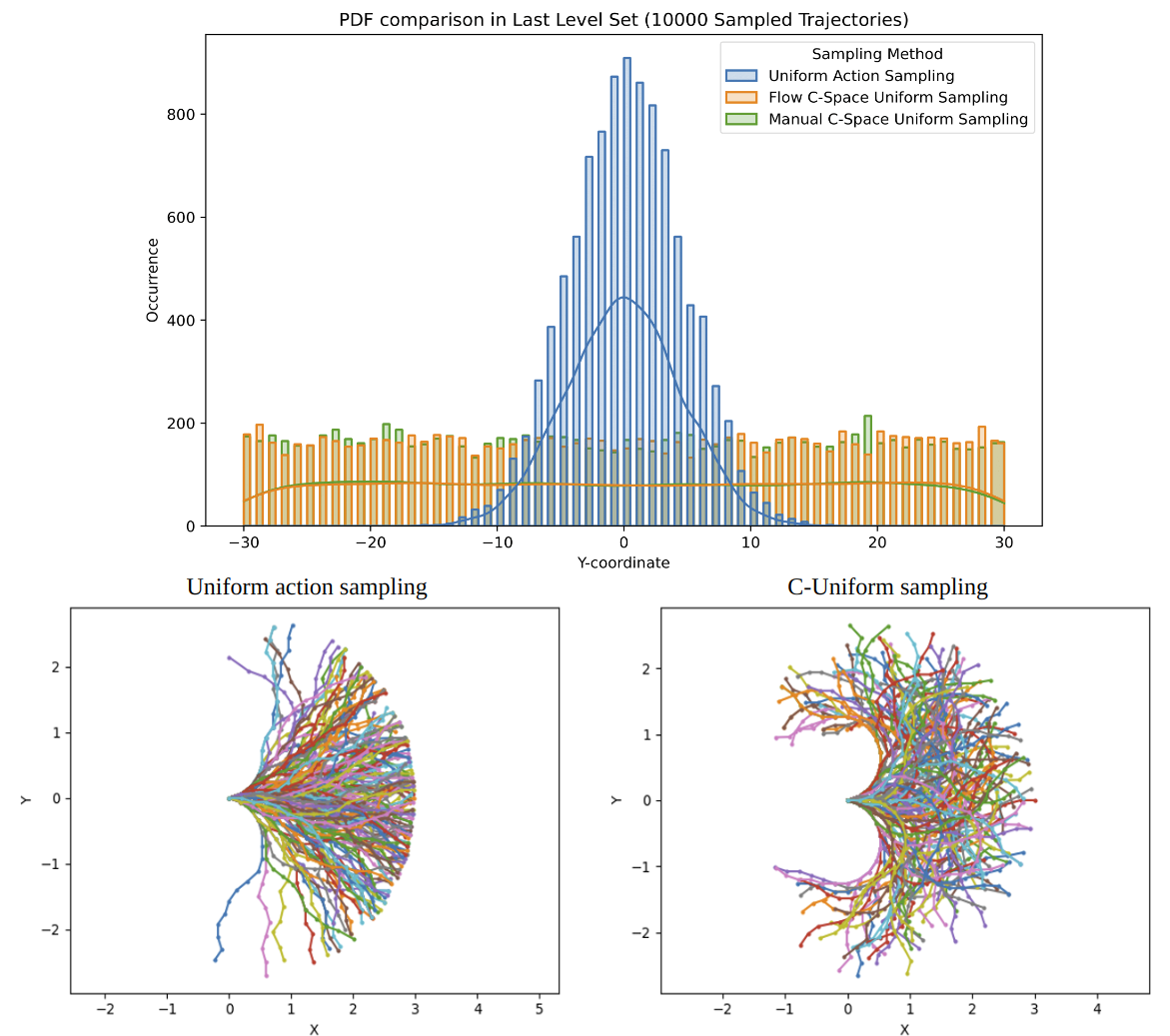}

        \label{fig:simple_system}
    \end{subfigure}
        \vspace{-20pt} % Adjust vertical space between subfigures

    \caption{Comparison of uniformly sampling the actions versus \ourmethod{} trajectories for the 1-D random walk system (top figure) and the Dubins car (bottom left vs. right). For the Dubins car, note how long sharp turns are not generated by uniformly sampling the control inputs (bottom left).}
    \label{fig:combined_figure}
\end{figure}

To make the case regarding sampling control inputs more concrete, consider the simple example of a robot on the line.
In Figure~\ref{fig:combined_figure}, we show the distribution of a robot which chooses between left, right and stay actions with equal probability. The blue distribution shows the distribution of the robot's location after 30 steps. It is well known that the probability distribution function (pdf) resulting from this random walk strategy is Bernoulli (it is Gaussian if the actions are chosen according to a Gaussian distribution). Clearly, the distribution is heavily weighted around the mean value and the likelihood of generating trajectories that visit points away from the mean value is extremely low. For example, an extreme trajectory which makes a sharp right turn (involving $k$ ``right" actions) is generated with the exponentially low probability of $3^{-k}$.
It is worth noting that such trajectories are not anomalies in motion planning and show up often
as part of optimal trajectories (e.g. for the Dubins' car) or from bang-bang controllers
and can be critical for optimality and safety. 

For motion planning, it is desirable to generate trajectories which  equally weigh all reachable configurations. In this paper, we introduce the notion of \emph{\ourmethody{}} where we seek to generate trajectories with the following property: Let $L((t)$ be the set of configurations reachable by the robot in time $t$. C-uniform trajectories are a collection of trajectories such that, for each $t$, configurations in $L(t)$ are sampled uniformly at random. Figure~\ref{fig:combined_figure} shows the distribution given by C-Uniform strategies generated by our method for two different systems: a one-dimensional random walker and a two-dimensional Dubins vehicle. We note that the notion of \ourmethody{} is different than uniformly sampling the configuration space where there is no notion of trajectory and the goal is to generate configuration samples which uniformly sample the C-Space.

It turns out that the need for C-Uniformity is widely acknowledged in the literature. As an example, consider the popular Model Predictive Path Integral (MPPI) method which starts with a set of sampled trajectories which are weighted according to a cost measure whose weighted sum yields the control input. In the original MPPI method~\cite{williams_information_2017}, the samples are generated using a Gaussian distribution over the control inputs. These trajectories are  clustered around a mean trajectory as described above. When approaching an obstacle, this centrality becomes a problem. As a result, the community proposed variations such as log-MPPI~\cite{mohamed_autonomous_2022} to flatten the trajectory distribution. More recently, Jia et. al~\cite{jia_towards_2024} proposed using entropy as a measure of trajectory diversity. Since the entropy is maximized when the distribution is uniform, this metric would favor uniformity.  In this paper, we formalize this intuition as~\ourmethody{} and make the following contributions:

\begin{enumerate}
    \item We formalize the notion of \ourmethody{} where a set of trajectories sample the level-sets of robot trajectories uniformly at random.
    \item We show how to compute \ourmethod{} trajectories: For the basic 1-D case, we derive a closed-form solution. For the general case, we present  optimization-based solution which is formulated as a network-flow optimization problem which can be solved in polynomial time to find weights for control inputs which lead to \ourmethody. 
    \item To show the utility of our algorithm, we introduce a new variant of the Stochastic Model Predictive Control (SMPC) solution based on our trajectory sampling strategy. We compare it against state-of-the-art variants and demonstrate its effectiveness both in simulations and on the F1Tenth racer platform.

\end{enumerate}

\section{Related Work} \label{sec:related_work}
Trajectory generation is a vast subject that lies at the heart of stochastic processes~\cite{kalakrishnan2011stomp}, and control theory~\cite{kazim2024recent}. A big portion of these studies is focused on understanding the behavior of a system whose evolution is governed by a differential equation. Generating control input or actions to generate trajectories with a desired property (e.g. optimality, safety) is studied in the context of motion planning and control. It turns out that controlling for the \emph{distribution} of trajectories over the entire C-space is relatively less studied. 

The most related work in the motion planning context is the uniform sampling of the C-Space~\cite{lavalle_planning_2006}. In this work, we are concerned with uniformly sampling the set of valid robot trajectories. To do so, we incrementally build either explicitly or implicitly through sampling, reachable level-sets of the robot. Level-set based methods~\cite{mitchell_comparing_2007} rely on the dynamic programming principle to generate all states which can be reached within a given time (or number steps) limit. They are heavily used for optimization-based motion planning and reinforcement learning. The dynamic programming principle (or the Hamilton-Jacobi equations) can be used to compute actions that are optimal in some sense. In this work, our goals is to generate actions which control the distribution across the entire level set. 

Calculus of variations~\cite{salamat2021control} provides an alternative to level-set/dynamic-programming based methods where we can differentiate a cost-functional to improve a trajectory iteratively. Such methods usually converge to local minima. 

In recent years, model predictive control (MPC) methods which combine the strengths of both approaches have become increasingly popular~\cite{hansen2022temporal}. In a nutshell, MPC methods solve an associated optimization problem to generate a trajectory, but rather than executing the entire solution, only a first step is executed. After this step, the control loop is closed by collecting measurements, updating states, and resolving the optimization problem. 

Our work is closely related to the Model Predictive Path Integral version of MPC, which, in essence, replaces the optimization step with an essentially exhaustive search-based strategy. More specifically, in the optimization stage, a dense sampling of trajectories is generated to compute the cost functional.

The original MPPI paper samples the input according to a Gaussian to generate the controllable distribution around a baseline trajectory~\cite{williams_information_2017}. Follow up extensions include, log-MPPI~\cite{mohamed_autonomous_2022}, which flattens the sampling distribution or input-lifting strategies~\cite{kim_smooth_2022} to smoothen the input sequence without using any external smoother.
Alternative approaches integrate external optimizations to guide MPPI trajectories into feasible regions. For example, Stein Variational Guided MPPI introduces mode-seeking optimizers to find optimal distribution peaks and shift trajectory generation towards those peaks~\cite{honda_stein_2024}. Similarly, Risk-Aware MPPI utilises CVaR optimization to produce safer trajectories~\cite{yin_risk-aware_2022}, while PRIEST uses projection techniques to ensure trajectories remain in feasible regions~\cite{rastgar_priest_2024}. Recent work combines MPPI and auxiliary controllers to identify the possible catastrophic cases and avoid them~\cite{trevisan_biased-mppi_2024}.  

In this paper, we leverage our results on \ourmethody{} and present a new MPPI variant which uses  \ourmethod{} sampling to generate candidate strategies and show that it significantly improves the performance of the resulting controller.

\section{Notation and Preliminaries} \label{sec:preliminaries}

We consider a robot system which evolves according to the following state space model:
\vspace{-4pt} % Reduce space before the equation
\begin{equation}
\statevec_{t+1} = \mathbf{F}(\statevec_t, \mathbf{u}_t)  
\label{eqn:dynamics}
\end{equation}
\vspace{-13pt}

where $\statevec \in \mathcal{X} \subseteq \mathcal{R}^p$, $\mathbf{u} \in \mathcal{U} \subseteq \mathcal{R}^q$  are the state and control of dimensions with $p$ and $q$, respectively. It is also assumed that the system has the Lipschitz continuity, meaning system responses to the state or input changes are bounded and predictable. This assumption is used to predict the forward propagation of the system. 
%\todotxt{figure out specifically what we want here}
Throughout the text, it is implicitly assumed that all control inputs are valid, i.e., $\mathbf{u} \in  \mathcal{U}$. 

A trajectory of length $T$ is specified by an  initial state $\statevec_0$ and control sequence $U =(\mathbf{u}_0, \mathbf{u}_1, \dots, \mathbf{u}_{T-1})$. The resulting trajectory is obtained by recursively applying $F$ with inputs $\statevec_i$ and $\mathbf{u}_i$. We use the shorthand notation $F(\statevec_0, U)$ to denote the state arrived after applying the input sequence $U$.

We define a \textit{Level Set} $L_t$ as the set of all states $\mathbf{x} \in \mathcal{X}$ such that there is a control sequence $U$ of length $t$ with $x = F(\statevec_0, U)$.

We use the Lebesgue measure to quantify the size of sets in  the C-space.  For a set \( A \) in the  Euclidean space, the \(d\)-dimensional Lebesgue measure is used to measure its \(d\)-dimensional ``volume". For example, 1, 2 or 3 dimensional Lebesgue measures correspond to length, area, and volume respectively. For our purposes, since we will be sampling the level-sets, it is helpful to consider each level-set to be divided into small, disjoint regions of measure $\delta$. 
Measurable sets are obtained by taking the unions of these regions. We assume that $\delta$ is small so that a Euclidean metric can be used as a measure.
For the rest of this paper, we denote the Lebesgue measure by \( \mu \) and refer to it simply as the measure. The uniform probability distribution assigns the probability $p(\Delta) = \mu(\Delta) / \mu(L)$ to each measurable region $\Delta$ (obtained by taking the unions of small $\delta$-regions. %$L$ is the level-set in consideration.

The probabilities associated with level sets are computed recursively as follows. 
Let $L = L_i$ and $L' = L_{i+1}$ be any two consecutive level-sets and suppose we are given.
Our goal is to eventually assign $p(x, u)$ for each $x \in L$ and $u \in \mathcal{U}$. When these are given, the probabilities can be computed in a straightforward manner:
\vspace{-3pt} % Reduce space before the equation
\begin{equation}
p(x') = \sum_{x \in L} \sum_{u: x'= F(x, u) } p(x, u)p(x)
\label{eqn:lprobs}
\end{equation}

\vspace{-3pt}
To make these definitions concrete, consider the one dimensional random walk with step size 1. After $t$ steps, the level set is the line segment $L = [-t, +t]$. Our goal is choose control inputs such that for any subset of $L$ of length $l$, the robot is in this subset with probability $l/L$.

We are now ready to state the main problem studied in this paper: 
\begin{problem}[\ourmethod-sampling]    
Given an initial state $\statevec_0$, a system model (Equation~\ref{eqn:dynamics}) choose control action probabilities $p(x,u)$ for each state such that the probability distribution associated with each level-set (Equation~\ref{eqn:lprobs} is uniform.) 
\end{problem}

%\section{Problem Formulation}  \label{sec:formulation}
%\input{sections/3-problem_formulation}

\section{Method} \label{sec:approach}
In this section, we present our approach to generate \ourmethod{} trajectories. We start with the one-dimensional basic random-walk and present an analytical solution. Next, we present our general optimization-based solution. 

\subsection{Solution for the one-dimensional case}

In this section, we consider the following simple version of the problem where the system evolves according to $\statevec_{t+1} = \statevec_t + u \Delta t$ with $u \in [-1,1]$. We take  $\Delta t = 1$ to simplify the notation. Let $u_i, i=1, \ldots 2k+1$ uniformly-sampled control inputs such that $|u_i u_{i+1}| < \delta$ where $\delta$ a desired level of resolution for \ourmethody.

Let $L$ and $L'$ be two consecutive level-sets with $n = \lceil \mu(L)/ \delta \rceil$ and $m = \lceil \mu(L') / \delta \rceil$ regions. In this specific case, $m = n + 2k$. We represent each region with a state that corresponds to its mid-point. 

The problem is now to assign $p(x_i, u_i)$ to each cell in $L$ and its control input to achieve uniformity. 

Consider the following control probabilities:
\begin{equation}
    p(x_i, u) = [n-i+1, 1, ...., 1, i]/m
\label{eqn:uniformcontrol}
\end{equation}

See Figure~\ref{fig:1d_distribution} for an example. In other words, the first (i.e. leftmost state in $L$ with $i=1$) goes left with probability $n/m$. The probability of the leftmost action reduces linearly to $1/m$  as the state moves from left to right whereas the probability of the rightmost action increases linearly. 
It can be verified that this probability assignment is a valid solution by checking:

(1) the control probabilities (the rows in Figure~\ref{fig:1d_distribution}) add up to 1: 
\begin{equation*}
    \sum_u p(x_i, u) = 1 \;  \forall x_i \in L \\
\end{equation*}
\vspace{-3pt}
and they yield a uniform distribution for the $L'$:
\begin{equation*}
    p(x') = \sum_{x\in L} \sum_{u: x + u = x'} p(x, u) p(x) = 1/m  \; \forall x' \in L'
\end{equation*}
That is, the columns in Figure~\ref{fig:1d_distribution} add up to $1/m$. 

Therefore, if the level set $L$ is \ourmethod, the control probabilities given in Equation~\ref{eqn:uniformcontrol} will lead to a \ourmethod{} distribution for $L'$. 
As a concrete example, Figure~\ref{fig:1d_distribution} shows the distribution with 
\(n=5\), \(m=5+2k=9\) where \(k=2\). In this table form, the rows correspond to states in $L$
and columns correspond to states in $L'$.
The total incoming flow to each of the \(5\) nodes in level set \(L\) sums to 9, and the total outgoing flow from each of the \(9\) nodes in level set \(L'\) sums to 5, hence uniformity is maintained.  Note that, to obtain the probabilities, these numbers should be divided by $9$.

\begin{figure}
    \captionsetup{aboveskip=0pt, belowskip=-10pt} % Set specific spacing for this figure
    \centering
    \begin{tikzpicture}[scale=0.38]
    \def\rows{5}
    \def\cols{9}

    \node[anchor=east] at (-0.05, -2.5) {$n=5$}; % "n=5" on the left side
    \node[anchor=south] at (4.5, 0.05) {$m=9$}; 
    \node[anchor=west] at (8.7, -5.5) {$sum$}; 
    
    \foreach \j in {1,...,9} {
        \node[anchor=center] at (\j-0.5, -5.5) {5}; % Bottom row values
    }
    \foreach \i in {1,...,5} {
        \node[anchor=center] at (9.5, -\i+0.5) {9}; % Right side column values
    }

    % Draw the table
    \foreach \i in {1,...,\rows} {
        \foreach \j in {1,...,\cols} {
            % Draw each cell
            \draw[thick] (\j-1,-\i+1) rectangle (\j,-\i); 
        }
    }
    
    \node at (0.5, -0.5) {5};    % (1,1)
    \node at (1.5, -0.5) {1};    % (1,2)
    \node at (2.5, -0.5) {1};    % (1,3)
    \node at (3.5, -0.5) {1};    % (1,4)
    \node at (4.5, -0.5) {1};    % (1,5)

    \node at (1.5, -1.5) {4};    % (2,2)
    \node at (2.5, -1.5) {1};    % (2,3)
    \node at (3.5, -1.5) {1};    % (2,4)
    \node at (4.5, -1.5) {1};    % (2,5)
    \node at (5.5, -1.5) {2};    % (2,6)

    \node at (2.5, -2.5) {3};    % (3,3)
    \node at (3.5, -2.5) {1};    % (3,4)
    \node at (4.5, -2.5) {1};    % (3,5)
    \node at (5.5, -2.5) {1};    % (3,6)
    \node at (6.5, -2.5) {3};    % (3,7)

    \node at (3.5, -3.5) {2};    % (4,4)
    \node at (4.5, -3.5) {1};    % (4,5)
    \node at (5.5, -3.5) {1};    % (4,6)
    \node at (6.5, -3.5) {1};    % (4,7)
    \node at (7.5, -3.5) {4};    % (4,8)

    \node at (4.5, -4.5) {1};    % (5,5)
    \node at (5.5, -4.5) {1};    % (5,6)
    \node at (6.5, -4.5) {1};    % (5,7)
    \node at (7.5, -4.5) {1};    % (5,8)
    \node at (8.5, -4.5) {5};    % (5,9)

    % \node at (10.5, -2.5) {\(\times \frac{1}{9}\)};
    \node[scale=1.5] at (11.5, -2.5) {\(\times \frac{1}{9}\)};

\end{tikzpicture}
\caption{\ourmethod{} control inputs the case of $n=5$ and $m=9$. Probabilities are obtained by dividing each entry by 9.\label{fig:1d_distribution}}
\end{figure}

\subsection{The general case}
\vspace{-16pt} % Reduce space between section heading and figure

\begin{figure}[!th]
    \captionsetup{aboveskip=-5pt, belowskip=-10pt} % Set specific spacing for this figure
    \begin{center}
        \includegraphics[width=1.0\columnwidth]{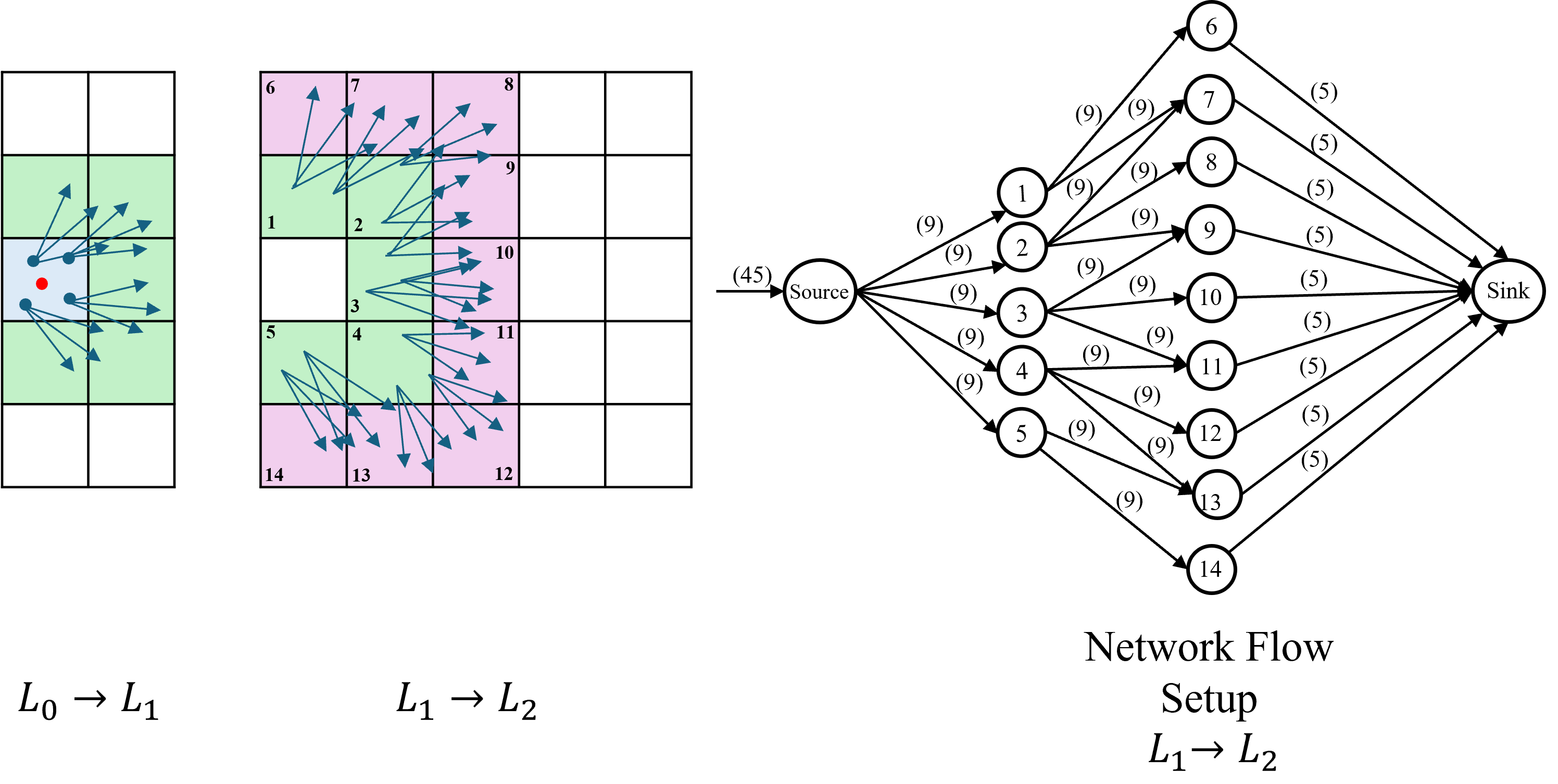}
    \end{center}
    \caption{Overview of approach: We generate level sets by densely sampling control inputs and tiling $\delta$-measure regions over each level set. To compute the action probabilities, a flow network is generated whose maximum-flow yields \ourmethody.}
    \label{fig:approach_general}
\end{figure}

We now consider the problem of generating \ourmethod{} control inputs for arbitrary systems. 
At a high-level, we iteratively compute \ourmethod{} control inputs one level set at a time. Suppose we have reached level set $t$ and are generating control inputs from level set $t$ to $t+1$. We will use $L$ and $L'$ to denote these level sets.  
First, we generate a tiling of both $L$ and $L'$ with disjoint $\delta$-measure cells and record control inputs, which take the robot from state $x \in L$ to $x' \in L'$.
The full algorithm is presented in Algorithm~\ref{algo:flow}.
To compute the probability for the control inputs, we create a flow network which has two layers corresponding to the cells in $L$ and $L'$, along with a source and a sink.  See Figure~\ref{fig:approach_general}. Similar to the previous section, let $n$ and $m$ denote the number of cells in these levels and $p = m \times n$.
From the source node, there is an arc to each of the $n$ nodes in the first layer. Each edge has capacity 1. The arcs from layer 1 to layer 2 correspond to control inputs: there is an arc from $x$ to $x'$ if there exists a control input $u$ such that $x' = F(x, u)$. These arcs have capacity $m$.
At the final layer, there is an arc from each node to the sink with capacity $n$. The sink has an outgoing node capacity of $p$. 

\vspace{-10pt}
\begin{algorithm}
\caption{\algoname} \label{algo:flow}
\KwIn{
    $\mathcal{C}$: Configuration space;\\
    $\delta$: Grid dimension for partitioning;\\
    $N$: Number of time steps in a trajectory;\\
    $\mathcal{U}$: Set of actions;\\
    $\Delta t$: Time step duration;
}
\KwOut{(grid, action probability distribution) pair;}

PartitionBoundedConfigurationSpace($\mathcal{C}$, \(\delta\))\;
$\mathcal{G} = \{\mathcal{G}[t]\}_{t=0}^{N}$ where $\mathcal{G}[t]$ is the set of reachable cells at time step $t$\;
\For{each time step $t = 0$ to $N-1$}{
    \(\mathcal{E} = []\)\;
    \For{each grid $g_t \in \mathcal{G}[t]$}{
        Sample $n$ points in $g_t$ to obtain $\{x_{t,i}\}_{i=1}^n$\;
        \For{each sampled point $x_{t,i} \in \{x_{t,i}\}_{i=1}^n$}{
        % \For{each sampled points \(x_t\)}{
            \For{each action $u \in \mathcal{U}$}{
            $x_{t+1}\leftarrow f(x_{t}, u, \Delta t)$\;
            Find \( g_{t+1} \in \mathcal{G}[t+1]\) s.t. \(x_{t+1} \in g_{t+1}\)\;
            $\mathcal{E} \gets \mathcal{E} \cup \{(g_t, g_{t+1}, u)\}$\;
        }
    }
    Solve $\text{MaxFlow}(\mathcal{G}[t], \mathcal{G}[t+1], \mathcal{E})$\;
    Compute $\mathrm{Prob}(g_t, u)$ based on $\text{flow}_{g_t, u}$\;
    }
}
\Return $(g, \mathrm{Prob}(g, u))$ pair\;
\end{algorithm}
\vspace{-10pt}

We can now establish a 1-1 correspondence between the maximum flow in this network and \ourmethody:

\begin{lemma}
    There exist \ourmethod{} control inputs between levels $L$ and $L'$ if and only if $n \times m$ units of flow can be transported by the flow network described above. 
\end{lemma}

The lemma can be easily verified in both directions: 
If $p = m n$ units of flow is feasible, this means that each of the $m$ arcs into the sink is at full capacity $n$ -- establishing uniformity. This in turn means that the $n$ arcs from the source node are also at full capacity $m$. Hence, each node in level $L$ must be pushing $m$ units of flow. The action probabilities $p(x, u)$  can be obtained by dividing the outgoing flow by $m$. Note that $\sum_u p(x, u) = 1$ since the total flow into the node is $m$. 

For the other direction, it can be verified that assigning flow values according to the \ourmethod{} probabilities leads to a flow of capacity $p$.

Figure~\ref{fig:combined_figure} shows \ourmethod{} strategies obtained by sampling according to the closed-form solution and the network-flow based solution. Both solutions achieve \ourmethody.

\section{Experiments} \label{sec:experiments}
In this section, we conduct experiments to demonstrate the overall sampling performance of our \ourmethod~sampling method compared with the baselines .
% to point out the key performance differences under different environmental settings. 
Specifically, we select MPPI~\cite{williams_information_2017} and log-MPPI\cite{mohamed_autonomous_2022} as our baselines for the following experiments since they focus on the distribution of the trajectory sampling to generate trajectory rollouts without external optimizers to modify the distribution.

\subsection{Experiment setup}

\textbf{Vehicle Model}. All following experiments are conducted with similar settings. We model the vehicle as a Dubin's car. The kinematic model of the vehicle is expressed as Eq.~\ref{eq;kinematic_model}. We denote $\statevec = [x,y, \theta] \in \mathbb{R}^3 $ as the vehicle's state and $u = \omega$ as the angular velocity.

\vspace{-22pt} % Reduce space before the equation
\begin{align}   \label{eq;kinematic_model}
    \dot{\statevec} &= f(\statevec, u) = 
    \begin{bmatrix}
        \dot{x} \\
        \dot{y} \\
        \dot{\theta}
    \end{bmatrix}
    =
    \begin{bmatrix}
        v\cos\theta \\
        v\sin\theta \\
        \omega
    \end{bmatrix}
\end{align}

\begin{table*}[htbp]
    \centering
    \caption{\small Coverage Performance Comparison of MPPI, log-MPPI, and \ourmethod~at Varying Sample Sizes}
    \begin{adjustbox}{max width=\textwidth}
    \begin{tabular}{c|ccc|ccc|c}
    \toprule
    \multirow{4}{*}{\textbf{\begin{tabular}[c]{@{}c@{}}Number of\\ Sampled \\ Trajectories\end{tabular}}} & \multicolumn{7}{c}{\textbf{Coverage Performance $\boldsymbol{\uparrow}$}} \\

    \cmidrule(lr){2-8} 
    & \multicolumn{3}{c|}{\textbf{MPPI}} & \multicolumn{3}{c|}{\textbf{log-MPPI}} & \multirow{2}{*}{\textbf{\ourmethod }} \\
    \cmidrule(lr){2-4} \cmidrule(lr){5-7}
     & \textbf{Low} & \textbf{Medium} & \textbf{High} & \textbf{Low} & \textbf{Medium} & \textbf{High} \\
    \midrule
    250  & 21 (0.09\%)    &  86 (0.36\%)  &  329 (1.37\%)    & 74 (0.31\%)     & 271 (1.13\%)    & 674 (2.81\%)    &   \textbf{737 (3.08\%)}  \\
    500  & 23 (0.10\%)   &  94 (0.39\%)   & 356 (1.48\%)    & 79 (0.33\%)   & 345 (1.44\%)     & 897 (3.74\%)    & \textbf{995 (4.15\%)}    \\
    1000 & 26 (0.11\%)    & 103 (0.43\%)    & 410 (1.71\%)    & 76 (0.32\%)   & 390 (1.63\%)    &  1140 (4.75\%)   & \textbf{1382 (5.83\%)}    \\
    2500 &  28 (0.12\%)   & 112 (0.47\%)    & 529 (2.20\%)    & 93 (0.39\%)    &  470 (1.96\%)    & 1420 (5.92\%)    & \textbf{1851 (7.71\%)}    \\
    5000 &  27 (0.11\%)   & 120 (0.50\%)    & 584 (2.43\%)    & 100 (0.42\%)    & 525 (2.19\%)    &  1637 (6.82\%)   & \textbf{2271 (9.46\%)}    \\
    10000 &  29 (0.12\%)   & 131 (0.55\%)    & 653 (2.72\%)    &  104 (0.43\%)   & 568 (2.37\%)    & 1838 (7.66\%)    & \textbf{2578 (10.74\%)}    \\
    \bottomrule
    \end{tabular}
    \end{adjustbox}
    \label{tab:comparison_coverage}
\end{table*}

\textbf{Parameters}. The general parameters for the following experiments are set as follows. The time horizon is $ T = 3s$, and the time discretization for the forward system propagation is $\Delta t = 0.2s$.  
We fix $\lambda=0.567$ as the temperature parameter for the MPPI methods and select different covariance values $\Sigma{u} = \{0.03, 0.1, 0.3\}$ for different exploration behaviors, labeling them as low, medium, and high in the tables, respectively.

% ---------------------------------
% The evaluation of the costs of trajectories uses the following cost formulation:
We define the costs of trajectories as Eq.~\ref{eq:cost}.
\begin{equation}    \label{eq:cost}
    \begin{split}
    &\ell(\statevec) =  \ell_{state}(\statevec) + \ell_{obs}(\statevec) \\
    &\ell_{state}(\statevec) = (\statevec_\text{p} - \statevec_{\text{goal}})^{T}(\statevec_\text{p} - \statevec_{\text{goal}})  \\
    &\ell_{obs}(\statevec) = 
        \begin{cases} 
            \textit{inf} \quad \text{if } \statevec \in \mathcal{O} \\
            0 \quad otherwise
        \end{cases}
    \end{split}
\end{equation}
We denote $\ell_{state}$ as the cost between the current state position $\statevec_p=(x,y)$ and the goal position $\statevec_{\text{goal}}=(x,y)$. We define the obstacle cost as $\ell_{obs}(\statevec)$, where $\mathcal{O} \in \mathcal{C}$ defines the obstacle in the configuration space. 

\textbf{Runtime and Space Complexity Analysis}. 
While our approach requires tiling $\delta$-measure cells in the C-Space, we do not need to explicitly create a grid for our approach. Instead, we compute each cell’s indices on demand by dividing state values by the grid size $\delta$.
The numbers of reachable cells in the resulting level sets in experiments were 1, 8, 33, 99, 247, 517, 974, 1703, 2820, 4427, 6646, 9622, 13530, 18613, 25107, 33184. The total time it takes to assign action probabilities for all level sets is about 14000 seconds, using the platform Ubuntu 20.04, i7-7700K, and GTX 1080.

The runtime complexity of the algorithm is primarily determined by the number of nodes and cubic runtime complexity for solving max flow problem. Approximately, $O\left( \sum_{t=0}^{T-1} n_t^3 \right)$, where $n_t$ is the number of nodes at each level set $t$. While the approach is computationally intensive, these probabilities can be computed in advance and used during execution without incurring any significant additional computational load than uniformly sampling the inputs.

\subsection{Coverage Analysis}
We measure how many of those grid cells were covered by the sampled trajectories from the baselines and \ourmethod. We report \textbf{coverage} as the percentage of the total reachable space. We summarize the comparison results for 2-second long trajectories under different settings in Table~\ref{tab:comparison_coverage}. The results indicate that \ourmethod~trajectories achieve up to 40\% more coverage than the best baseline.

\subsection{Environment with an Obstacle}
\vspace{-8pt} % Reduce space between section heading and figure
\begin{figure}[thp]
    \captionsetup{aboveskip=3pt, belowskip=0pt} 
    \begin{center}
        \includegraphics[width=1.0\columnwidth]{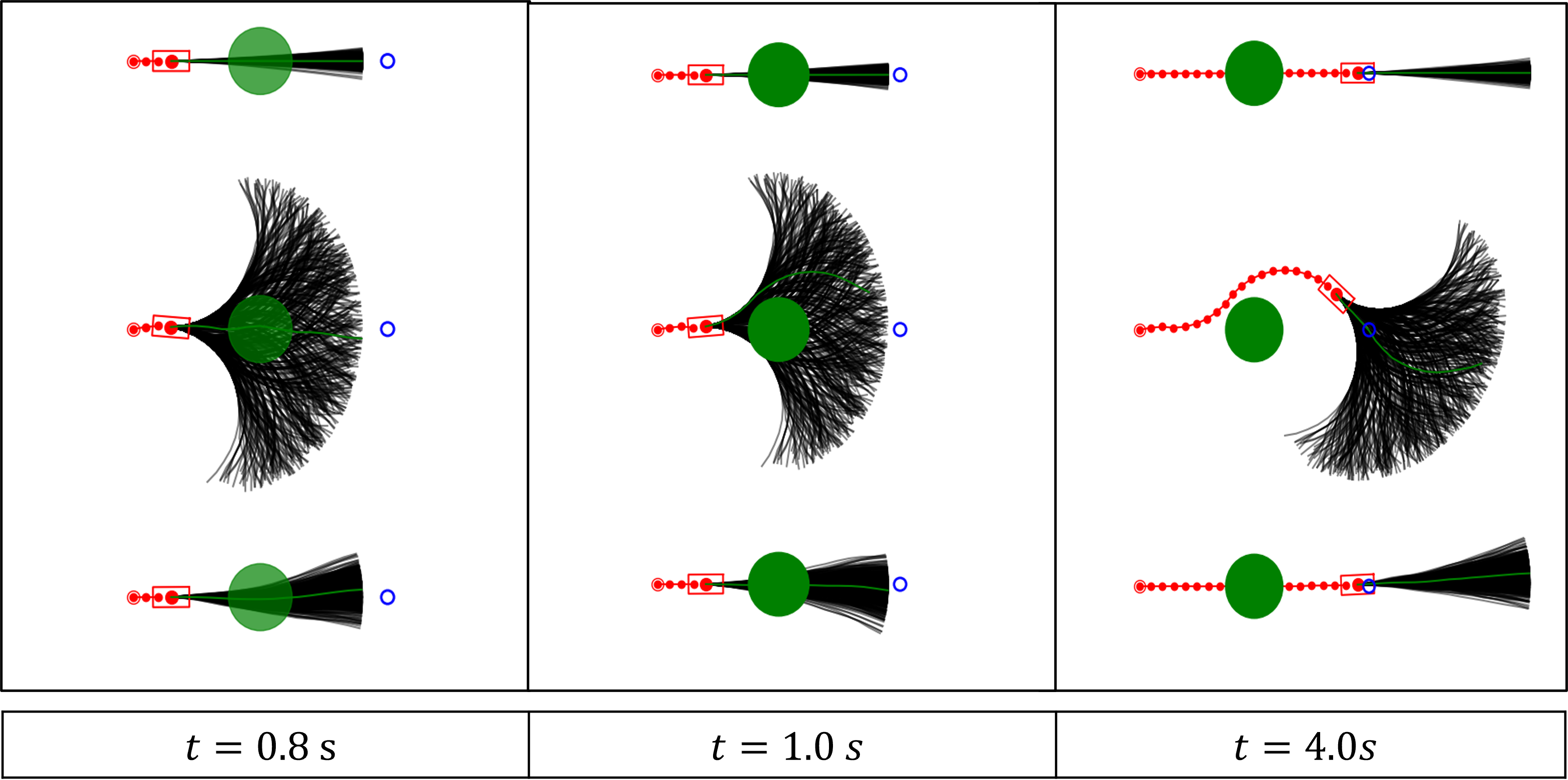}
    \end{center}
    \caption{Comparison of the MPPI, log-MPPI and \ourmethod{} for the suddenly appearing obstacle experiment. The obstacle becomes visible after 1.0s of the robot movement (middle). The baselines can not avoid collisions because their samples lie inside the obstacle. In contrast, \ourmethod{} samples  avoid the obstacle by achieving a wider coverage, which allows for successful re-planning (right).}
    \label{fig:sudden_obs}
\end{figure}

In this experiment setting, we design an environment where a circular obstacle will be visible after some time the vehicle begins its movement. The obstacle position is also selected uniformly in the interval $[-r_{obs}, r_{obs}]$, where $r_{obs}$ is the radius of the obstacle. The initial and goal positions remain the same across all methods. The number of sampled trajectories is set to $N = 500$ for every method. Fig.~\ref{fig:sudden_obs} shows qualitative results of the superior adaptability of the \ourmethod{} as the baselines failed to generate a feasible no-collision trajectory.

In addition, Table~\ref{tab:comparison} compares the baselines and our \ourmethod{} sampling method under different obstacle appearance times. The performance metric is the \textbf{success rate} ($SR$) of the generated trajectories, where being successful means that the method was able to generate no-collision trajectory between the initial point and the goal location. For each appearance time, the experiment contains 20 different obstacle positions. The success rate was found by dividing the number of successful runs by the total number of runs.

\begin{table}[thp]
    \centering
    \caption{\small Success Rates for Varying Obstacle Appearance Times}
    \begin{adjustbox}{max width=\columnwidth}
    \begin{tabular}{c|c|c|c|c|c|c}
    
    \toprule
    \multicolumn{7}{c}{\textbf{Success Rate (\%) $\boldsymbol{\uparrow}$}} \\
    \midrule
    \textbf{Methods} & \multicolumn{2}{c|}{\textbf{MPPI}} & \multicolumn{2}{c|}{\textbf{log-MPPI}} & \multicolumn{2}{c}{\textbf{\ourmethod}}\\
   \midrule 
   % \cmidrule(lr){2-3} \cmidrule(lr){4-5} \cmidrule(lr){6-7}
    \diagbox[width=10em, height=5em]{\textbf{Appearance} \\ \textbf{Time (s)}}{\textbf{No. of} \\ \textbf{Trajectories}} & \textbf{500} & \textbf{1000} & \textbf{500} & \textbf{1000} & \textbf{500} &  \textbf{1000}\\
    \midrule
    Fully Visible  & 1.0   & 1.0    & 1.0   & 1.0    & 1.0          &   1.0          \\
    0.2            & 1.0   & 1.0    & 1.0   & 1.0    & 1.0          &   1.0          \\
    0.5            & 0.55  & 0.85   & 0.95  & 0.95   & \textbf{1.0} &   \textbf{1.0} \\
    0.8            & 0.3   & 0.35   & 0.45  & 0.40   & \textbf{0.7} &   \textbf{0.65}\\
    1.0            & 0.05  & 0.15   & 0.15  & 0.25   & \textbf{0.25}&   0.25         \\
    \bottomrule
    \end{tabular}
    \end{adjustbox}
    \label{tab:comparison}
\end{table} 
\vspace{-9pt}
\subsection{Cluttered Environments}

In this experiment setting, we conduct experiments to evaluate the performance of \ourmethod~compared to two baselines in cluttered and complex environments. Each environment had multiple obstacles and a specific goal position. We randomly pick 10 initial positions for each environment. Similarly, we use \textbf{Success rate} ($SR$) as an evaluation metric to assess the performance. 
 
\begin{figure}[!htbp]
    \captionsetup{aboveskip=3pt, belowskip=-2pt} 
    \centering
    \begin{subfigure}[t]{0.9\columnwidth}  % Adjust the width for both figures
        \centering
        \includegraphics[width=0.9\columnwidth]{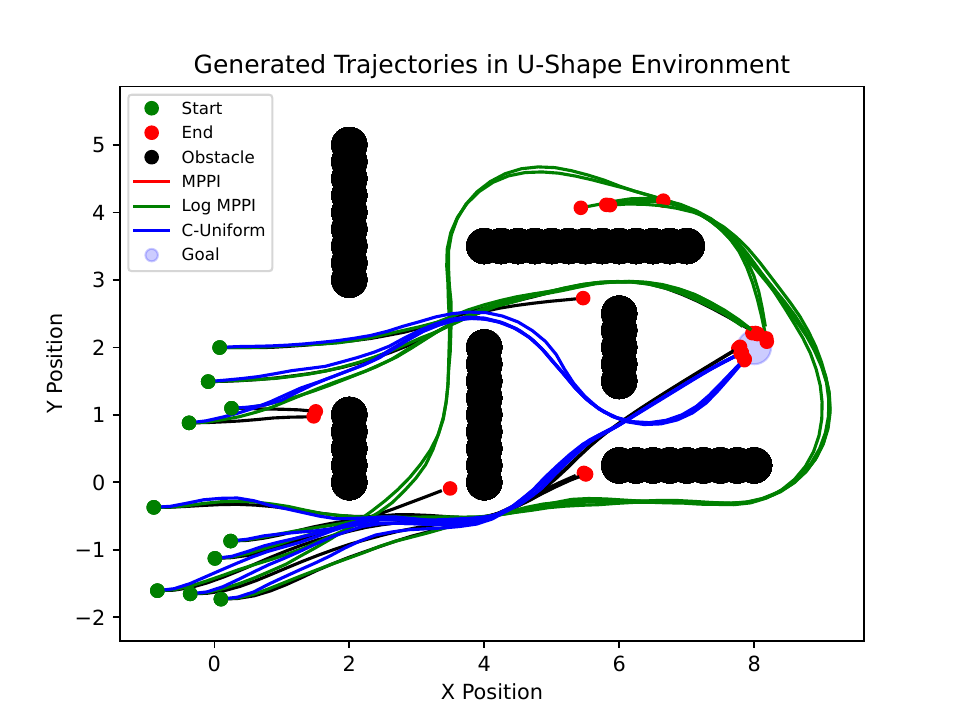}
        \caption{Environment with Rectangular Obstacles}
    \end{subfigure}%
    % \hfill
    \vspace{0pt}  % Adjust this value to control the vertical space between subfigures

    \begin{subfigure}[t]{0.9\columnwidth}  % Adjust the width for both figures
        \captionsetup{aboveskip=0pt, belowskip=0pt} 
        \centering
        \includegraphics[width=0.9\columnwidth]{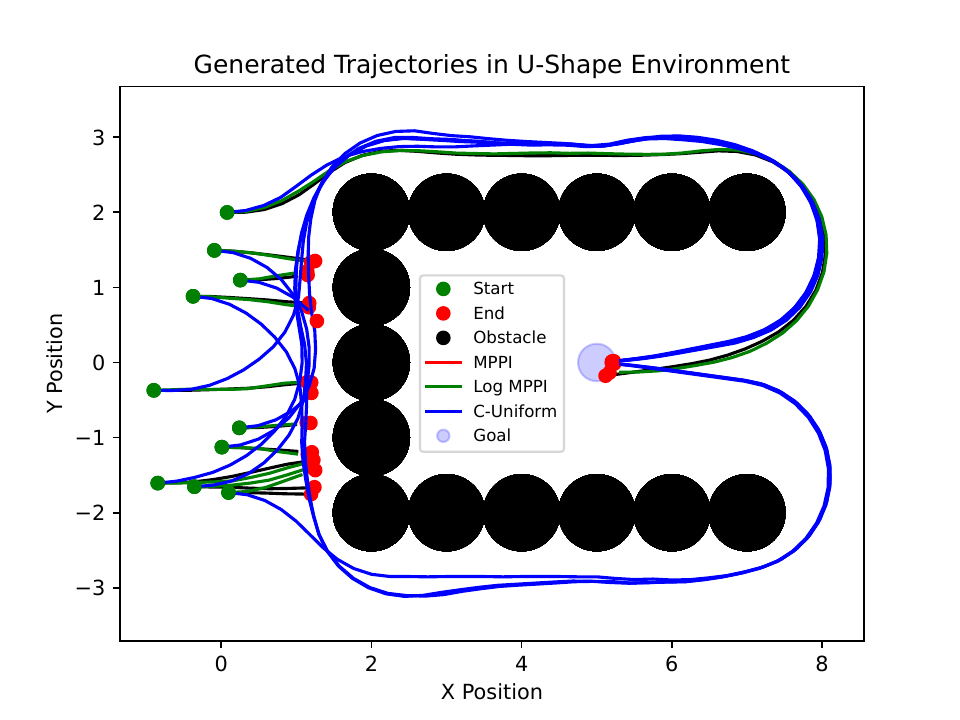}
        \caption{Environment with a U-Shaped Obstacle with a Goal Point Inside of the U}
    \end{subfigure}
    \captionsetup{aboveskip=0pt, belowskip=-10pt} 
    \caption{Performance comparison of MPPI, log-MPPI and \ourmethod{} with 1000  number of sampled trajectories in complex enviroments.}
    \label{comparison_cluttered}
\end{figure}

One run was considered successful if the vehicle completed the task by reaching the goal with no collisions. Based on the environment size we set a time limit: if the vehicle cannot reach to the goal position within the time limit, we consider the run as a failure. Table~\ref{tab:complex_sr_comparision} presents each method's success rate results with various sampled trajectories for two environments. The generated trajectories with 1000 sampled trajectories can be seen in Fig.~\ref{comparison_cluttered}. 

\begin{table}[thp]
    \centering
    \caption{\small Success Rates for Various Complex Environments}
    \begin{adjustbox}{max width=\columnwidth}
    \begin{tabular}{c|c|c|c|c|c|c}
    
    \toprule
    \multicolumn{7}{c}{\textbf{Success Rate (\%) $\boldsymbol{\uparrow}$}} \\
    \midrule
    \textbf{Methods} & \multicolumn{2}{c|}{\textbf{MPPI}} & \multicolumn{2}{c|}{\textbf{log-MPPI}} & \multicolumn{2}{c}{\textbf{\ourmethod}}\\
   \midrule 
   % \cmidrule(lr){2-3} \cmidrule(lr){4-5} \cmidrule(lr){6-7}
    \diagbox[width=10em, height=5em]{\textbf{No. of} \\ \textbf{Trajectories}}{\textbf{Environments}} & \textbf{U-Shaped} & \textbf{Rectangular} & \textbf{U-Shaped} & \textbf{Rectangular} & \textbf{U-Shaped} &  \textbf{Rectangular}\\
    \midrule
    500  & 0.1 (1)   & 0.1 (1)    & 0.1 (1)   & 0.4 (1)    & \textbf{0.8 (8)}  &   \textbf{1.0 (10)}         \\
    1000            & 0.1 (1)   & 0.3 (3)    & 0.1 (1)   & 0.6 (6)    & \textbf{0.8 (8)}          &   \textbf{1.0 (10)}          \\
    2500            & 0.1 (1)  & 0.3 (3)   & 0.1 (1)  & 0.8 (8)   & \textbf{0.8 (8)} &   \textbf{1.0 (10)} \\

    \bottomrule
    \end{tabular}
    \end{adjustbox}
    \label{tab:complex_sr_comparision}
\end{table}

The quantitative results show that our method has the highest success rate for every scenario in these settings. Log-MPPI also increased its performance for the rectangular environment.
On the contrary, the success rate for the U-shaped environment remained the same with the change in the number of sampled trajectories. 
In contrast to the non-exploitative nature of the baselines, our method performs the best in these cluttered environments.
Lastly, the results also show that uniformity in the configuration space helps explore the environment and leads to a solution, in this case, to the goal position. 

\subsection{Real Experiments}
\begin{figure}[thp]
    \captionsetup{aboveskip=-2pt, belowskip=-2pt} 
    \begin{center}
        \includegraphics[width=0.4\columnwidth]{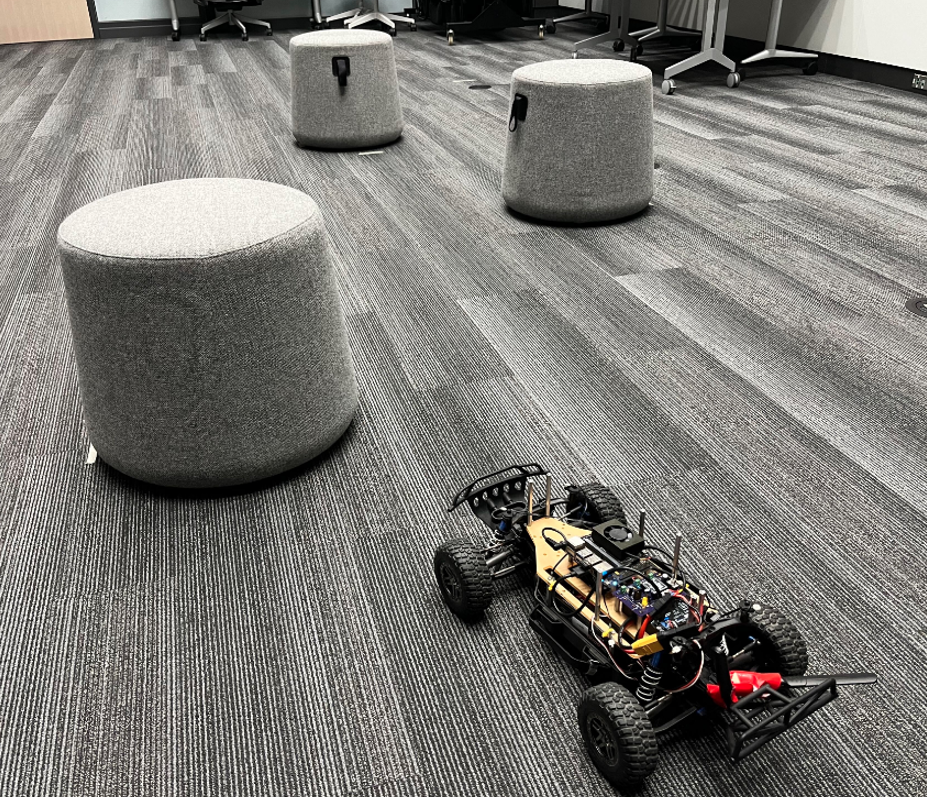}
        \includegraphics[width=0.7\columnwidth]{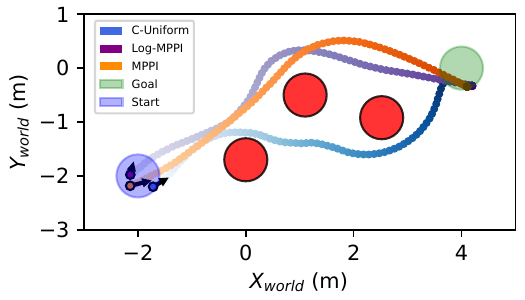}

    \end{center}
    \caption{\textbf{Top}: Experiment setup \textbf{Bottom}: Successful run trajectories for each method: MPPI, Log-MPPI, and \ourmethod}
    \label{fig:real_experiment}
\end{figure}

We tested our algorithm and compared it against baselines on the F1Tenth racer platform~\cite{o2020f1tenth} in an environment with three circular obstacles (Figure~\ref{fig:real_experiment}). Phasespace X2E LED motion capture system was used for robot localization. 
We conducted motion planning trials where we fixed the goal position at location $(4m, 0m)$ in the workspace. We sampled initial locations from a small circle centered at $(-2m, -2m)$. The heading was chosen uniformly at random. For each method (MPPI, Log-MPPI and \ourmethod), we ran 5 trials. Figure~\ref{fig:real_experiment} shows a representative from each method.

Overall, Log-MPPI and \ourmethod{} methods reached the goal position in all trials. In contrast, MPPI had one unsuccessful run, resulting in an obstacle collision. In addition, we also noticed that \ourmethod{} was able to find narrow paths better than the baselines MPPI and Log-MPPI. Even though the average distances traveled by the vehicle with each method were pretty similar, MPPI: $7.63m$, Log-MPPI: $7.34m$, \ourmethod: $7.67m$, \ourmethod{} had the shortest distances in two specific runs, with $7.03m$ and $6.93m$, respectively.  

\section{Conclusion and Future Work} \label{sec:conclusion}
In this paper, we sought to generate random trajectories which uniformly sample the configuration space. For this purpose, we introduced the notion of \ourmethody{} where at each time step $k$, the robot's probability of being at a configuration in the $k^{th}$ level set is uniform. We showed how to generate control input probabilities to achieve \ourmethody{} by formulating a max-flow problem. We then introduced a new version of the Model Predictive Path Integral controller which uses \ourmethod{} control inputs to generate proposal trajectories. 
Simulation and real experiments showed that \ourmethody{} yields superior controllers in terms of success rate in environments with obstacles. 

One of the drawbacks of our method is that it is computationally expensive. However, this is not a big detriment for real-time operation since the control input probabilities can be computed in advance. Furthermore, parallel processing techniques can be utilized to speed up the computation. Our immediate next step is to demonstrate real-time performance in realistic environments. We will also work on further improving sample efficiency by incorporating the environment map and/or the sensor footprints into the computation of control input probabilities.

\section{Acknowledgement}
We thank Qingyuan Jiang and Burak Mert Gonultas for the helpful discussions and editing. 

% \section*{APPENDIX}
% % \input{sections/appendix}

% \section*{ACKNOWLEDGMENT}
% \input{sections/acknowledgment}

% \clearpage
\bibliographystyle{IEEEtran}
% \bibliography{main.bib}

\begin{thebibliography}{10}
\providecommand{\url}[1]{#1}
\csname url@rmstyle\endcsname
\providecommand{\newblock}{\relax}
\providecommand{\bibinfo}[2]{#2}
\providecommand\BIBentrySTDinterwordspacing{\spaceskip=0pt\relax}
\providecommand\BIBentryALTinterwordstretchfactor{4}
\providecommand\BIBentryALTinterwordspacing{\spaceskip=\fontdimen2\font plus
\BIBentryALTinterwordstretchfactor\fontdimen3\font minus \fontdimen4\font\relax}
\providecommand\BIBforeignlanguage[2]{{%
\expandafter\ifx\csname l@#1\endcsname\relax
\typeout{** WARNING: IEEEtran.bst: No hyphenation pattern has been}%
\typeout{** loaded for the language `#1'. Using the pattern for}%
\typeout{** the default language instead.}%
\else
\language=\csname l@#1\endcsname
\fi
#2}}

\bibitem{orthey_sampling-based_2024}
\BIBentryALTinterwordspacing
A.~Orthey, C.~Chamzas, and L.~E. Kavraki, ``\BIBforeignlanguage{en}{Sampling-{Based} {Motion} {Planning}: {A} {Comparative} {Review}},'' \emph{\BIBforeignlanguage{en}{Annual Review of Control, Robotics, and Autonomous Systems}}, vol.~7, no.~1, pp. 285--310, July 2024. [Online]. Available: \url{https://www.annualreviews.org/content/journals/10.1146/annurev-control-061623-094742}
\BIBentrySTDinterwordspacing

\bibitem{williams_information_2017-1}
\BIBentryALTinterwordspacing
G.~Williams, N.~Wagener, B.~Goldfain, P.~Drews, J.~M. Rehg, B.~Boots, and E.~A. Theodorou, ``Information theoretic {MPC} for model-based reinforcement learning,'' in \emph{2017 {IEEE} {International} {Conference} on {Robotics} and {Automation} ({ICRA})}, May 2017, pp. 1714--1721. [Online]. Available: \url{https://ieeexplore.ieee.org/document/7989202}
\BIBentrySTDinterwordspacing

\bibitem{ota_trajectory_2019}
\BIBentryALTinterwordspacing
K.~Ota, D.~K. Jha, T.~Oiki, M.~Miura, T.~Nammoto, D.~Nikovski, and T.~Mariyama, ``Trajectory {Optimization} for {Unknown} {Constrained} {Systems} using {Reinforcement} {Learning},'' in \emph{2019 {IEEE}/{RSJ} {International} {Conference} on {Intelligent} {Robots} and {Systems} ({IROS})}, Nov. 2019, pp. 3487--3494, iSSN: 2153-0866. [Online]. Available: \url{https://ieeexplore.ieee.org/abstract/document/8968010}
\BIBentrySTDinterwordspacing

\bibitem{kavraki1996probabilistic}
L.~E. Kavraki, P.~Svestka, J.-C. Latombe, and M.~H. Overmars, ``Probabilistic roadmaps for path planning in high-dimensional configuration spaces,'' \emph{IEEE transactions on Robotics and Automation}, vol.~12, no.~4, pp. 566--580, 1996.

\bibitem{karaman_sampling-based_2011}
\BIBentryALTinterwordspacing
S.~Karaman and E.~Frazzoli, ``\BIBforeignlanguage{en}{Sampling-based algorithms for optimal motion planning},'' \emph{\BIBforeignlanguage{en}{The International Journal of Robotics Research}}, vol.~30, no.~7, pp. 846--894, June 2011, publisher: SAGE Publications Ltd STM. [Online]. Available: \url{https://doi.org/10.1177/0278364911406761}
\BIBentrySTDinterwordspacing

\bibitem{williams_information_2017}
\BIBentryALTinterwordspacing
G.~Williams, P.~Drews, B.~Goldfain, J.~M. Rehg, and E.~A. Theodorou, ``Information {Theoretic} {Model} {Predictive} {Control}: {Theory} and {Applications} to {Autonomous} {Driving},'' July 2017, arXiv:1707.02342 [cs]. [Online]. Available: \url{http://arxiv.org/abs/1707.02342}
\BIBentrySTDinterwordspacing

\bibitem{mohamed_autonomous_2022}
\BIBentryALTinterwordspacing
I.~S. Mohamed, K.~Yin, and L.~Liu, ``\BIBforeignlanguage{en}{Autonomous {Navigation} of {AGVs} in {Unknown} {Cluttered} {Environments}: log-{MPPI} {Control} {Strategy}},'' July 2022, arXiv:2203.16599 [cs, eess]. [Online]. Available: \url{http://arxiv.org/abs/2203.16599}
\BIBentrySTDinterwordspacing

\bibitem{jia_towards_2024}
\BIBentryALTinterwordspacing
X.~Jia, D.~Blessing, X.~Jiang, M.~Reuss, A.~Donat, R.~Lioutikov, and G.~Neumann, ``Towards {Diverse} {Behaviors}: {A} {Benchmark} for {Imitation} {Learning} with {Human} {Demonstrations},'' Feb. 2024, arXiv:2402.14606 [cs]. [Online]. Available: \url{http://arxiv.org/abs/2402.14606}
\BIBentrySTDinterwordspacing

\bibitem{kalakrishnan2011stomp}
M.~Kalakrishnan, S.~Chitta, E.~Theodorou, P.~Pastor, and S.~Schaal, ``Stomp: Stochastic trajectory optimization for motion planning,'' in \emph{2011 IEEE international conference on robotics and automation}.\hskip 1em plus 0.5em minus 0.4em\relax IEEE, 2011, pp. 4569--4574.

\bibitem{kazim2024recent}
M.~Kazim, J.~Hong, M.-G. Kim, and K.-K.~K. Kim, ``Recent advances in path integral control for trajectory optimization: An overview in theoretical and algorithmic perspectives,'' \emph{Annual Reviews in Control}, vol.~57, p. 100931, 2024.

\bibitem{lavalle_planning_2006}
S.~M. LaValle, \emph{Planning algorithms}.\hskip 1em plus 0.5em minus 0.4em\relax Cambridge university press, 2006.

\bibitem{mitchell_comparing_2007}
I.~Mitchell, ``Comparing {Forward} and {Backward} {Reachability} as {Tools} for {Safety} {Analysis},'' Apr. 2007, pp. 428--443.

\bibitem{salamat2021control}
B.~Salamat, N.~A. Letizia, and A.~M. Tonello, ``Control based motion planning exploiting calculus of variations and rational functions: A formal approach,'' \emph{IEEE Access}, vol.~9, pp. 121\,716--121\,727, 2021.

\bibitem{hansen2022temporal}
N.~Hansen, X.~Wang, and H.~Su, ``Temporal difference learning for model predictive control,'' \emph{arXiv preprint arXiv:2203.04955}, 2022.

\bibitem{kim_smooth_2022}
\BIBentryALTinterwordspacing
T.~Kim, G.~Park, K.~Kwak, J.~Bae, and W.~Lee, ``Smooth {Model} {Predictive} {Path} {Integral} {Control} without {Smoothing},'' \emph{IEEE Robotics and Automation Letters}, vol.~7, no.~4, pp. 10\,406--10\,413, Oct. 2022, arXiv:2112.09988 [cs]. [Online]. Available: \url{http://arxiv.org/abs/2112.09988}
\BIBentrySTDinterwordspacing

\bibitem{honda_stein_2024}
\BIBentryALTinterwordspacing
K.~Honda, N.~Akai, K.~Suzuki, M.~Aoki, H.~Hosogaya, H.~Okuda, and T.~Suzuki, ``Stein {Variational} {Guided} {Model} {Predictive} {Path} {Integral} {Control}: {Proposal} and {Experiments} with {Fast} {Maneuvering} {Vehicles},'' Feb. 2024, arXiv:2309.11040 [cs, math]. [Online]. Available: \url{http://arxiv.org/abs/2309.11040}
\BIBentrySTDinterwordspacing

\bibitem{yin_risk-aware_2022}
\BIBentryALTinterwordspacing
J.~Yin, Z.~Zhang, and P.~Tsiotras, ``Risk-{Aware} {Model} {Predictive} {Path} {Integral} {Control} {Using} {Conditional} {Value}-at-{Risk},'' Sept. 2022, arXiv:2209.12842 [cs, eess]. [Online]. Available: \url{http://arxiv.org/abs/2209.12842}
\BIBentrySTDinterwordspacing

\bibitem{rastgar_priest_2024}
\BIBentryALTinterwordspacing
F.~Rastgar, H.~Masnavi, B.~Sharma, A.~Aabloo, J.~Swevers, and A.~K. Singh, ``{PRIEST}: {Projection} {Guided} {Sampling}-{Based} {Optimization} for {Autonomous} {Navigation},'' \emph{IEEE Robotics and Automation Letters}, vol.~9, no.~3, pp. 2630--2637, Mar. 2024, conference Name: IEEE Robotics and Automation Letters. [Online]. Available: \url{https://ieeexplore.ieee.org/document/10412189/?arnumber=10412189}
\BIBentrySTDinterwordspacing

\bibitem{trevisan_biased-mppi_2024}
\BIBentryALTinterwordspacing
E.~Trevisan and J.~Alonso-Mora, ``Biased-{MPPI}: {Informing} {Sampling}-{Based} {Model} {Predictive} {Control} by {Fusing} {Ancillary} {Controllers},'' \emph{IEEE Robotics and Automation Letters}, vol.~9, no.~6, pp. 5871--5878, June 2024, arXiv:2401.09241 [cs]. [Online]. Available: \url{http://arxiv.org/abs/2401.09241}
\BIBentrySTDinterwordspacing

\bibitem{o2020f1tenth}
M.~O'Kelly, H.~Zheng, D.~Karthik, and R.~Mangharam, ``F1tenth: An open-source evaluation environment for continuous control and reinforcement learning,'' \emph{Proceedings of Machine Learning Research}, vol. 123, 2020.

\end{thebibliography}

\end{document}